# A Fusion of Labeled-Grid Shape Descriptors with Weighted Ranking Algorithm for Shapes Recognition

[1]Jamil Ahmad, [1]Zahoor Jan, [1]Zia-ud-Din and [2]Shoaib Muhammad Khan

[1]Department of Computer Sciences, Islamia College Peshawar (Chartered University)
Peshawar, P.O. Box 25000, Khyber Pakhtunkhwa, Pakistan
[2]Department of Computer Science, National University of Computer and Emerging
Sciences Peshawar Campus, Pakistan

**Abstract:** Retrieving similar images from a large dataset based on the image content has been a very active research area and is a very challenging task. Studies have shown that retrieving similar images based on their shape is a very effective method. For this purpose a large number of methods exist in literature. The combination of more than one feature has also been investigated for this purpose and has shown promising results. In this paper a fusion based shapes recognition method has been proposed. A set of local boundary based and region based features are derived from the labeled grid based representation of the shape and are combined with a few global shape features to produce a composite shape descriptor. This composite shape descriptor is then used in a weighted ranking algorithm to find similarities among shapes from a large dataset. The experimental analysis has shown that the proposed method is powerful enough to discriminate the geometrically similar shapes from the non-similar ones.

## INTRODUCTION

Shape is considered as a basic characteristic to describe visual content. It serves as the basic feature for object recognition. Shapes description has its applications in robotics, fingerprint analysis, face recognition, automatic target recognition, document analysis (OCR), handwriting mapping and image retrieval etc.

In the field of computer vision, object recognition has been achieved by matching the properties of object shapes extracted from the object image. Instead of matching objects directly, their geometric and statistical properties are measured which are merely a set of numbers to describe the essential geometric characteristics of a shape called descriptors and then these descriptors are compared together to find the similarities and differences among various shapes. In the literature, there exist two ways for extracting these features from objects. In region based methods the entire set of pixels that makeup the object are considered. Whereas in contour based approach only the boundary pixels are taken into account for feature extraction. Both have their strengths and weaknesses. For instance region based methods are considered to be more suitable for general applications and are robust to noise and distortions [1]. Contour based methods have been widely used but the contour usually is affected greatly by noise and slight distortions may cause errors in matching.

A large number of global and local shape descriptors have been introduced in the past which have been evaluated on certain publicly available shape databases including the MPEG-7 Part-B and Kimia shape databases [2, 3]. Some features have been extracted from shapes directly in the space domain while others have used the various available transform domains [4-7] to extract features in order to represent the shapes and then match them.

---

**Corresponding Author:** Jamil Ahmad, Department of Computer Sciences, Islamia College Peshawar Khyber Pakhtunkhwa, Pakistan. Tel: +92-91-9216948.

**Shapes Recognition Is a Three-step Process:**

**Feature Extraction:** Involves extraction of the shape descriptor either by considering its contour or interior pixels.

**Similarity Measurement:** Involves matching the shape descriptors from different shapes in order to determine similarity or dissimilarity. Numerous techniques exist for similarity measurement and are described in detail by [8]

**Recognition:** Refers to the problem of determining the category of an unknown shape. This is usually done using various classification techniques.

Some descriptors focus on capturing local shape features, they tend to fail to capture global characteristics and vice versa. Hence such shape features often fail to accurately classify shapes within the same class when they have different contour signature because of difference of their local properties. The objective of this work is to extract multiple features from a shape that are derived from both the shape's interior and contour and are invariant to geometric transformations and slight distortions at the edges. To achieve such characteristics with shape features, they have been derived using a labeled grid based approximation of the shape. The following sections describe how the labeled-grid is calculated, how the shape features are extracted from the labeled grid representation of the shape and how the matching and ranking of similar shapes occur.

**Related Work:** Previous researches have shown that recognizing objects using their shape features is a very powerful and effective approach. Numerous techniques exist for extracting features from shapes that can be used to represent and match shapes. Some of the earlier features designed were shape signature, signature histogram, shape context, shape invariant moments, shape matrix, curvature, spectral features etc. Every technique that exists has some pros and cons. The complexity and variety of content of images makes it impossible for any single descriptor to be appropriate for all kinds of shapes. Therefore fusion of multiple shape features has been investigated. This section introduces a very few efforts carried out in pursuit of developing methods for shapes recognition and fusion of shape features.

Belongie *et al* [9] proposed a method for shape representation called shape context. Shape context is a contour based descriptor that represents a shape using its boundary points. The distance of all boundary points of a shape are calculated with respect to any single boundary point. Objects are recognized by matching their shape contexts with previously known information regarding shape contexts.

The BSM descriptor represents a shape as equal sized regions (cells) in the form of a grid[10]. The shape itself is represented as a set of key points. The spatial relations among the key points from neighbour regions are computed. Each cell receives votes from its neighbors. This contribution depends upon the distance of the point from the cell centroid [11]. Each cell is represented as a bin in the feature histogram. To achieve rotation invariance, the shape is rotated using hoteling transform. The BSM values are normalized which results in scale invariance. The classification of shapes uses an Error Correcting Output Code (ECOC) approach which operates by assigning a code word to each class in the dataset. The BSM has been applied on 70 classes of MPEG-7 dataset and 17 classes of gray level symbols from real environments. The accuracy of the BSM descriptor was recorded to be 74%.

The circular BSM is an improvement in previously discussed BSM[12]. It captures the significant shape characteristics in a correlogram structure by considering their spatial arrangement. The CBSM is rotational invariant by definition. Blurring (i.e. taking into consideration the neighboring cells) makes it tolerant to deformations. In order to make the descriptor rotation invariant, the shape is aligned to x-axis by taking the main diagonal that maximizes the sum of descriptor values as a reference point. For the classification, the ECOC framework is used. It is a meta-learning algorithm that operates by dividing a multiclass problem into a set of binary problems. Solves them individually and then aggregates their result to get the final result [13]. The CBSM improves the performance of BSM descriptor by increasing the accuracy from 74% to 78%.

Another method called co-Transduction is fusion based. In this method different similarity measures are fused together. It makes use of well-known shape features like Shape Context[9], Inner-Distance Shape Context [14] and DDGM [15] along with the proposed co-transduction approach to improve the performance of shape retrieval. The co-transduction approach makes use of two similarity measures for each query shape [16]. The algorithm extracts the most similar shapes from a large database into a pool for the other measure which is then used to do a re-ranking[17]. With co-transduction the bull's eye test was run with 97.7 % accuracy.

Mohammad Ali *et al*. [18] proposed that combining descriptors extracted using contour based and region based methods can improve retrieval performance. The main role in this paper was learning how to map the samples in high-dimensional observation space into the new manifold space so that the geometrically closer vectors belong to near semantics. The authors combined features that were calculated from the shapes considering different aspects and showed that the fusion of features brought considerable improvements to the recognition process.

Hamid *et al*. [19] used shape skeleton for feature extraction and represented those skeletons as connectivity graphs. Radius function is used along the skeletal curve segments to capture the geometric shape features. The topologies and geometric features of the connectivity graphs give us the feature that is used for measuring similarity or dissimilarity among shapes. For clustering and classification purposes a hierarchical clustering method is used. This method makes use of the distance measure calculated previously. Furthermore, for each class, a median skeleton is calculated and is used as the indicator of its related class. The resulted hierarchy of the shapes classes and their indicators are used for shape recognition. For classification of each query shape, the hierarchy is traversed in a top-down manner and comparing indicators as we move down the hierarchy. This technique is able to recognize shapes accurately up to 95%.

**The Proposed Labeled-grid Based Shape Features:** The proposed shape recognition method is based on fusion of structural and statistical shape features derived from both the shape contour and interior region from a labeled grid based shape representation. Efforts have been made to make the proposed feature invariant to translation, scale and rotation transformations. The proposed feature is based on grid based shape approximation so this makes it tolerant to slight deformations along the shape contour. The shape is divided into a square sized grid of size N x N. The size of N is taken to be odd. The shape is divided into grids in such a way that the central cell always overlaps onto the shape centroid. Aligning the grid with the shape's orientation makes the feature invariant to rotation. Computation of the proposed Grid-Based representation is explained as follows.

**Feature Computation**

**Obtaining Labeled-grid Based Feature:**

- Divide the shape into N x N cells where N is odd.
- For each cell of the grid calculate the percentage area covered by the shape pixels.
- Use this percentage to label the grid cells as
- Interior region (if most of the cell is occupied by the shape)
- Boundary region (if some portion of the cell is occupied by the shape pixels) and
- Background region

This process yields a square matrix which contains the shape labels as Interior, Boundary and Background. For the shape features only the interior and boundary labels are considered.

Then the grid-based shape feature is computed as the probability of occurrence of the interior and boundary regions away from the central grid cell in the form of rectangular shaped tracks shown in Figure 2. The grid cell labeled as 'C' is the central cell. The ones labeled as '1' is 'track-1' given that it is at distance 1 from C and the cells labeled as '2' gives us 'track-2' which is at distance 2 from the grid cell C. For a grid of size N x N, the number of such tracks will be $\frac{N-1}{2}$.

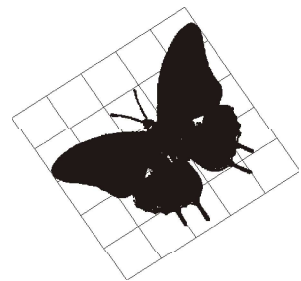

Fig. 1: The square shaped grid aligned with the shape orientation

| | 2 | 2 | 2 | 2 | 2 | |
| | 2 | 1 | 1 | 1 | 2 | |
| | 2 | 1 | C | 1 | 2 | |
| | 2 | 1 | 1 | 1 | 2 | |
| | 2 | 2 | 2 | 2 | 2 | |
| | | | | | | |

Fig. 2: The labeled shape matrix

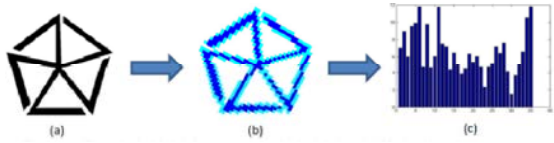

Fig. 3: (a) shape (b) the labeled shape matrix showing both the interior (blue) and boundary regions (cyan) (c) the centroid distance feature extracted from the labeled shape matrix

- The probability of the occurrence of interior and boundary regions in these tracks constitutes the shape feature.

The probabilities are computed as

$$P_d = \frac{n_t}{n_s} \qquad (1)$$

Where Pd is the probability of occurrence of interior or boundary region at distance d, nt is the number of interior or boundary regions and ns is the total number of interior or boundary regions in the shape.

This yields a shape descriptor derived from both the shape interior and boundary pixels.

**Grid Based Contour Shape Signature:** Contour shape descriptors are usually computed directly from the shape contour. Deformation along the shape contour greatly affects the descriptor so instead of calculating the feature from contour pixels directly it has been calculated using the labeled grid. Doing so will make our feature tolerant to slight deformations along the shape contour. Considering only the boundary regions from the labeled shape matrix extracted above, the centroid distance function (CDF) [20] which is basically the distances of the contour points at various angles from the shape centroid has been derived as shown in Figure 3. It is also used along the feature extracted above as structural representation of the shape under study.

**Building the Composite Shape Feature:** The two shape features extracted from the labeled shape matrix are strictly local so it was analyzed to use these features in conjunction with certain global statistical features and the results turned out to be very promising [21]. So a certain number of these global features including Eccentricity, Circularity, Aspect Ratio, Extent and solidity were also fused together with the features derived earlier to form a composite representation of shapes as shown in the Figure 4.

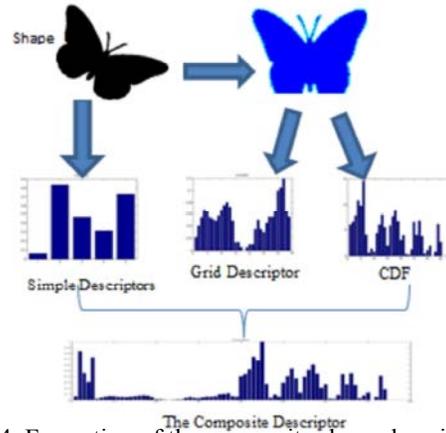

Fig. 4: Formation of the composite shape descriptor

Table 1: The grid-based feature vector.

| Tracks | Probability (Interior Region) | Probability (Boundary Region) |
|---|---|---|
| 1 | F1 | F1 |
| 2 | F2 | F2 |
| 3 | F3 | F3 |
| 4 | F4 | F4 |
| 5 | F5 | F5 |
| ... | ... | ... |

**Shape Matching:** In order to match shapes together a feature matching formula has been proposed. For two shapes P and Q, the similarity score in their features is given by

$$Sim(P,Q) = 1 - \frac{1}{n}\sum_{i=1}^{n} \frac{|F_{pi} - F_{Qi}|}{\max\left(F_{pi}, F_{Qi}\right)} \qquad (2)$$

Where Sim is the similarity score between shapes P and Q, FPi and FQi are the ith features of shape P and Q respectively, n is the total number of feature points. For shapes retrieval from the database the similarity scores are computed for each of the three features (Grid Descriptor, CDF and Simple Descriptors) individually using the given formula and then passed onto the weighted ranking algorithm for final ranking.

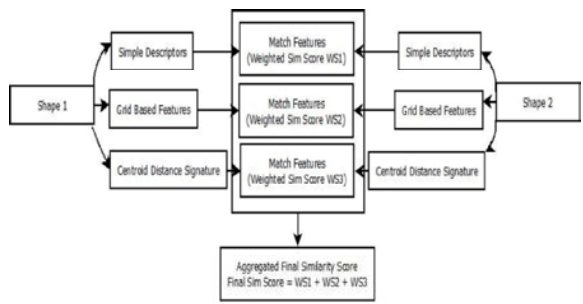

Fig. 5: Working of the weighted ranking algorithm

**Weighted Ranking Algorithm:** The weighted ranking algorithm works by assigning weights to individual shape features and then aggregates them into a final similarity score as illustrated in the figure below. The intermediate similarity scores obtained from individual comparison of features are aggregated using their designated weights to obtain the final similarity score between the two shapes.

**The Algorithm Works as Follows:**

- For shapes s1 & s2, calculate three different features i.e. Simple Descriptors, Grid based feature and centroid shape feature
- Find individual similarity scores among s1 & s2 using the proposed matching formula by matching these features individually
- Assign separate weights to each of these individual similarity scores and then accumulate them to find the final similarity score between shapes s1 & s2.

The shapes in a database are compared this way with the query shape and then are sorted in descending order based on their final rankings to produce the output. This weighted ranking algorithm fuses different features together for the purpose of shapes matching and retrieval has proven to be very flexible and has provided much better results as is evident from the retrieval results shown in figure 7.

**RESULTS**

**MPEG-7 Shapes Database:** For the purpose of evaluating the performance of the proposed algorithm, the MPEG-7 set B shape database has been used[22]. This database consists of 1400 shapes classified into 70 classes with 20 shapes in each class. Set B is used to test for similarity-based retrieval performance and to test the shape descriptors for robustness to various arbitrary shape distortions that include rotation, scaling, arbitrary skew, stretching, defection and indentation. This database is widely used for testing the performance of shape retrieval algorithms[2]. A sample of the shapes used from this database is shown in Figure 6.

**Performance Evaluation:** The proposed algorithm was implemented in MATLAB 7.14. The retrieval performance of the proposed algorithm is shown in figure 4. The first column shows the query image and the subsequent columns show the rank-wise retrieval of similar shapes using the proposed algorithm. The retrieval performance was tested using different combinations of weights in the proposed weighted ranking algorithm and the best set of weights was selected to run the given queries in Fig 7.

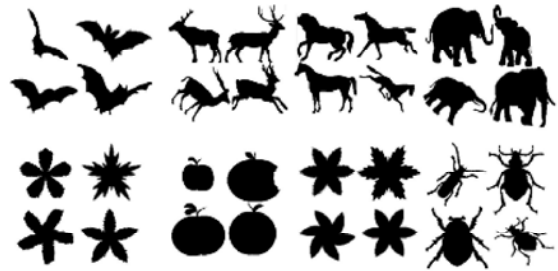

Fig. 6: Sample shapes from the mpeg-7 shapes database

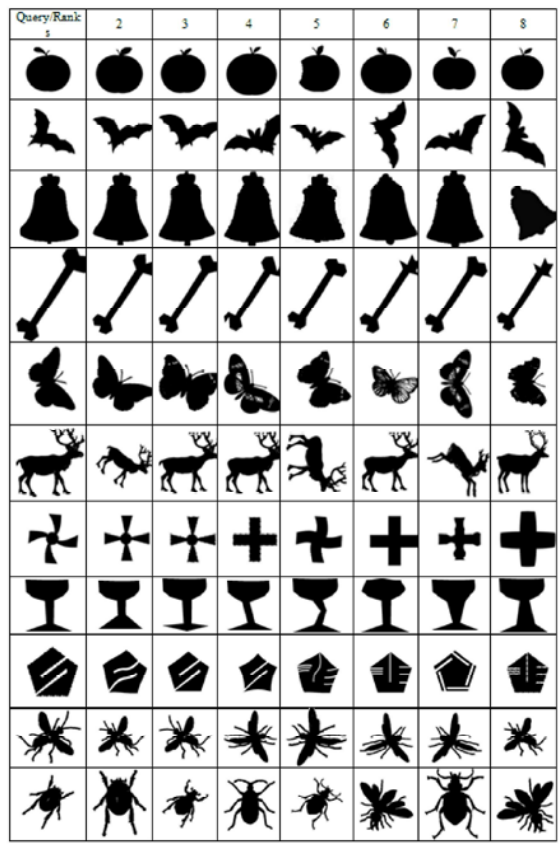

Fig. 7: Retrieval results for some random queries using the proposed algorithm

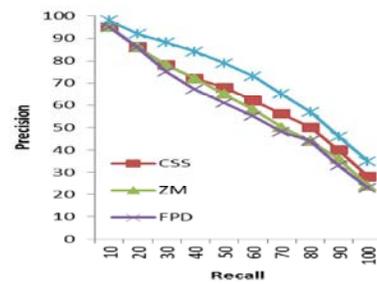

Fig. 8: Precision and recall performance comparison

Table 2: precision rates for low and high recall

| Method | Low Recall<br>Avg precision for<br>recall rates<50% | High Recall<br>Avg precision for<br>recall rates>50% |
|---|---|---|
| Proposed | 81.77 % | 50.63 % |
| FPD | 81.16 % | 49.15 % |
| ZM | 80.88 % | 43.94 % |
| CSS | 78.62 % | 41.81 % |

Table 3: retrieval rates (bull's eye test)

| Algo | Shape<br>Contexts<br>[9] | Distance<br>Set<br>[26] | Skeletal<br>Context<br>[27] | Inner<br>Distance<br>[14] | Proposed |
|---|---|---|---|---|---|
| Score | 76.51% | 78.38% | 79.92% | 85.40% | 86.65% |

The performance of the proposed algorithm was compared with other techniques "Zernike Moments"[23], CSS Descriptor [22] and "Farthest Point Descriptor"[24]. This analysis proved that the proposed algorithm has outperformed these methods.

The precision and recall are the most commonly used measures of retrieval performance. These methods are widely used across the literature to measure performance. These two measures are defined as:

$$Precision = \frac{No\ of\ relevant\ images\ retrieved}{Total\ No\ of\ images\ retrieved} \qquad (3)$$

$$Recall = \frac{No\ of\ relevant\ images\ retrieved}{Total\ Number\ of\ relevant\ images\ in\ the\ database} \qquad (4)$$

The average precision rates of low and high recall for the proposed, CSS, ZM and FPD using set-B are given in the table given below.

The retrieval performance of an algorithm is measured by the so-called bull's eye score[25]. Every shape in the database is compared to all other shapes and the number of shapes from the same class among the 40 most similar shapes is reported. The bulls-eye retrieval rate is the ratio of the total number of shapes from the same class to the highest possible number (which is 20 x 1,400). Thus, the best possible rate is 100 percent. From the retrieval rates collected in Table3, we can clearly see that our method has made a significant progress on this database.

## CONCLUSIONS

This paper presents fusion of labeled-grid based features that provides an efficient and flexible way to represent binary shapes and to effectively recognize them using a weighted ranking algorithm. The experimental results have shown that the proposed algorithm is very effective in shapes retrieval from a large database of binary shapes with variations in their size, orientation and shape. The results have also shown that the proposed algorithm is powerful enough to extract the geometrically similar shapes from the database.